\documentclass[sigconf]{acmart}

\title{Attention-Based Vandalism Detection in OpenStreetMap}

\setcopyright{none}

\usepackage[utf8]{inputenc}
\renewcommand\footnotetextcopyrightpermission[1]{}

\usepackage{hyperref}
\usepackage{subcaption}
\usepackage{xspace}
\usepackage{booktabs}
\usepackage{multirow}
\usepackage{makecell}
\usepackage{subcaption}
\usepackage{float}
\newcommand{\approach}{\textsc{Ovid}\xspace}

\newcommand\blfootnote[1]{%
  \begingroup
  \renewcommand\thefootnote{}\footnote{#1}%
  \addtocounter{footnote}{-1}%
  \endgroup
}

 \author{Nicolas Tempelmeier}
 \affiliation{%
   \institution{L3S Research Center\\Leibniz Universit\"at Hannover}
   \city{Hannover}
   \country{Germany}
 } \email{tempelmeier@L3S.de}

 \author{Elena Demidova}
 \affiliation{%
   \institution{\mbox{Data Science \& Intelligent Systems (DSIS)}\\ University of Bonn}
   \city{Bonn}
   \country{Germany}}
 \email{elena.demidova@cs.uni-bonn.de}



\begin{document}

\begin{abstract}
OpenStreetMap (OSM), a collaborative, crowdsourced Web map, is a unique source of openly available worldwide map data, increasingly adopted in Web applications.
Vandalism detection is a critical task to support trust and maintain OSM transparency. This task is remarkably challenging due to the large scale of the dataset, the sheer number of contributors, various vandalism forms, and the lack of annotated data.
This paper presents \approach{} - a novel attention-based method for vandalism detection in OSM. 
\approach{} relies on a novel neural architecture that adopts a multi-head attention mechanism to summarize information indicating vandalism from OSM changesets effectively.
To facilitate automated vandalism detection, we introduce a set of original features that capture changeset, user, and edit information.
Furthermore, we extract a dataset of real-world vandalism incidents from the OSM edit history for the first time and provide this dataset as open data.
Our evaluation conducted on real-world vandalism data demonstrates the effectiveness of \approach{}. 
\end{abstract}

\maketitle
\pagestyle{plain}
\blfootnote{\textcopyright Nicolas Tempelmeier, Elena Demidova 2022. This is the author's version of the work. It is posted here for your
personal use. Not for redistribution. The definitive version was published in the proceedings of The Webconference 2022 \url{https://doi.org/10.1145/3485447.3512224}.\\}

\section{Introduction}
\label{sec:intro}

\begin{figure*}
    \centering
    \begin{subfigure}[t]{0.245\textwidth}
        \includegraphics[width=\textwidth]{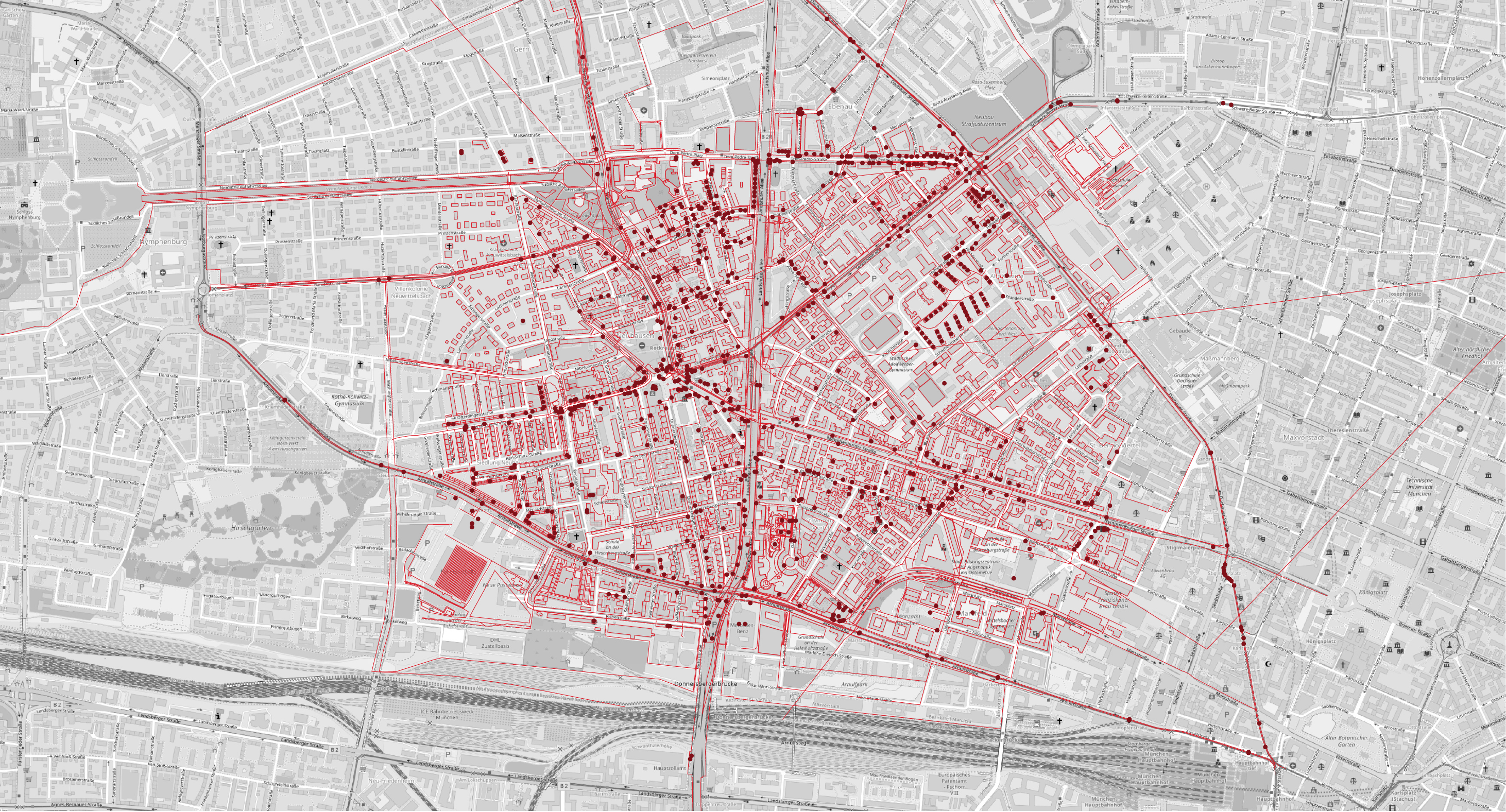}
        \caption{Deletion of large fractions of Munich}
    \end{subfigure}
    \begin{subfigure}[t]{0.245\textwidth}
        \includegraphics[width=\textwidth]{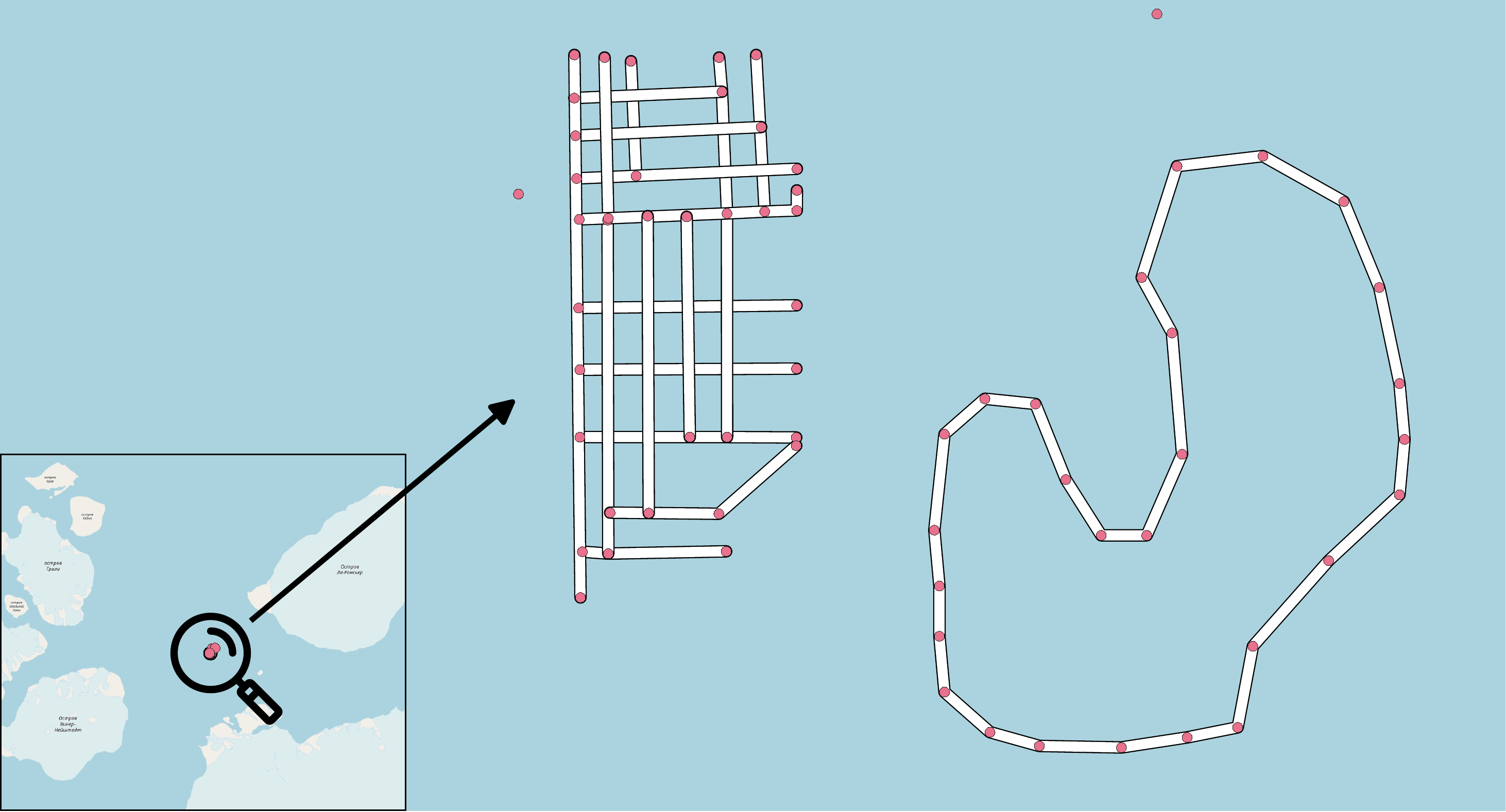}
        \caption{A fake town within the ocean}
    \end{subfigure}
    \begin{subfigure}[t]{0.245\textwidth}
        \centering
        \includegraphics[width=\textwidth]{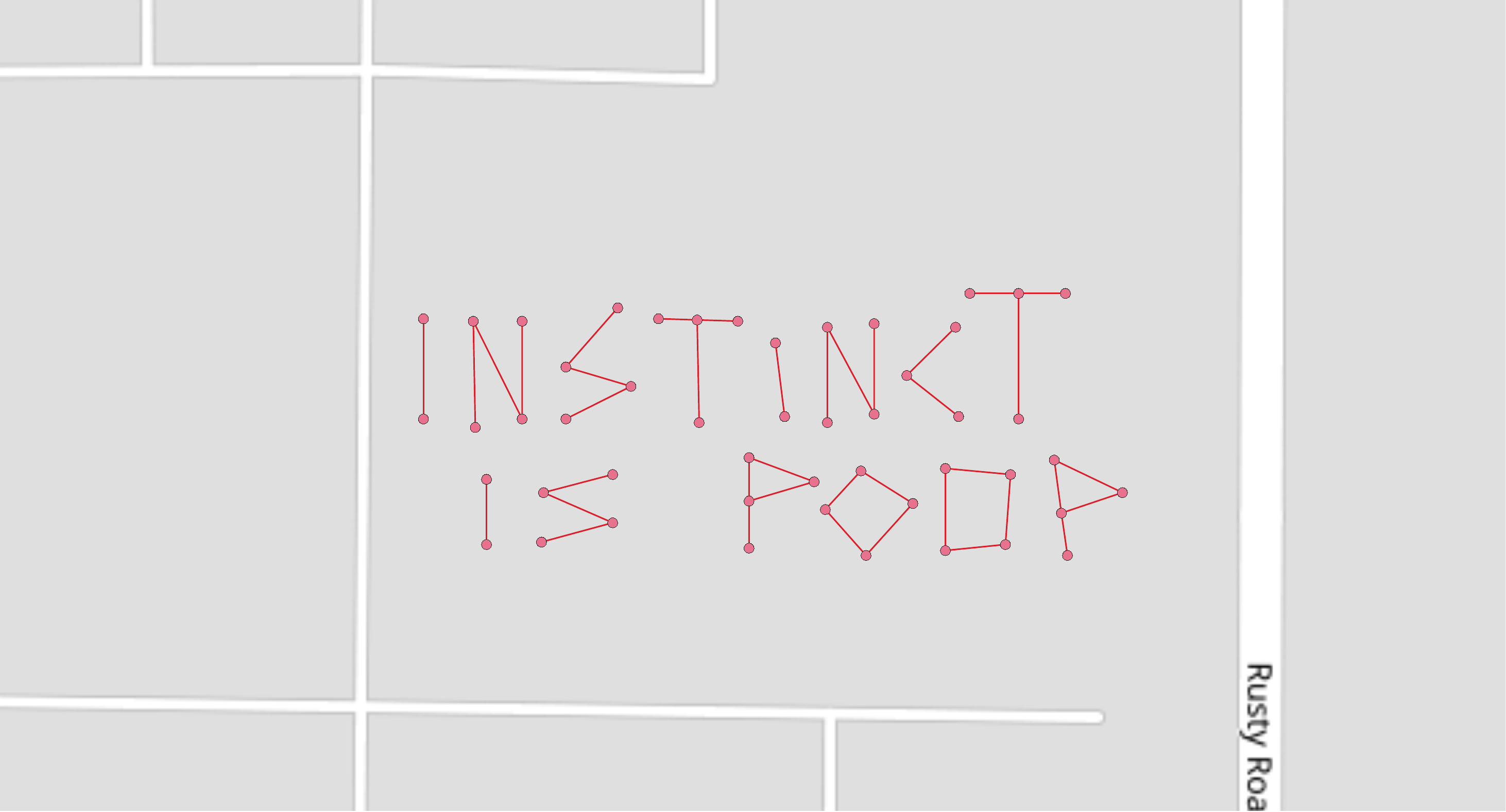}
        \caption{``Graffiti'' on the map \\} 
    \end{subfigure}
    \begin{subfigure}[t]{0.245\textwidth}
        \includegraphics[width=\textwidth]{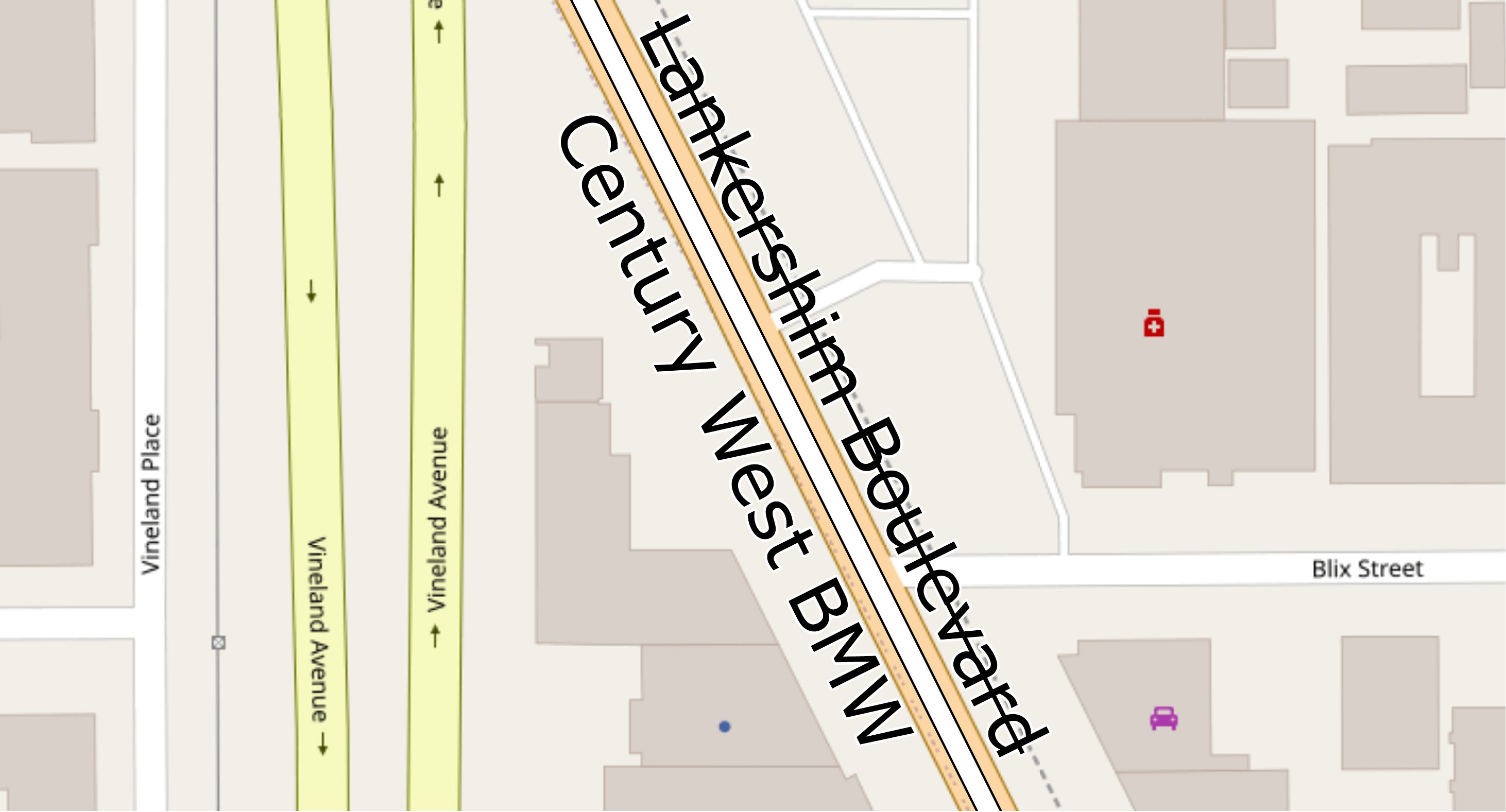}
         
        \caption{Road name replaced by an advertisement} 
    \end{subfigure}
    \caption{Real-world examples of different vandalism forms in OSM. Map data: \textcopyright OpenStreetMap contributors, ODbL.}
     \Description[Four map images]{Map images of a deleted city, a city within the ocean, a geometry forming letters, and a road with a replaced name}
    \label{fig:intro_example}
\end{figure*}

%

OpenStreetMap\footnote{\url{https://www.openstreetmap.org/}} 
(OSM) is a collaborative, crowdsourced Web map that evolved as the critical source of openly available volunteered geographic information \cite{JokarArsanjani2015, 10.1145/3459637.3482023, 10.1145/3459637.3482004}.
Like Wikipedia aiming to create an open Web encyclopedia, the OSM project aims to create a free and editable online map.
The amount of geospatial information in OSM is continuously growing. For instance, the number of nodes in OSM increased from $5.9 \cdot 10^9$ in March 2020 to $6.7 \cdot 10^9$ in March 2021.
With the OSM growth, quality assurance becomes essential to maintain the trustworthiness and accountability of OSM.
In OSM, contributors voluntarily provide geographic information and can add, modify or delete any objects captured by OSM. This openness makes the data particularly susceptible to vandalism.
%
Recently, the problem of vandalism detection in OSM has attracted interest of researchers \cite{ijgi9090504,9119753, vand_statemap, vand_cikm} and OSM contributors\footnote{\url{https://wiki.openstreetmap.org/wiki/Vandalism}}.
For example, the OSM community identified several vandalism cases in the context of the location-based game Pokémon Go \cite{ijgi9040197}, in which users added wrong information to the map to gain a game advantage.
Today, OSM provides data for various real-world applications, including knowledge graph refinement \cite{TEMPELMEIER2021349,10.1007/978-3-030-88361-4_4} and geographic information systems (GIS).
Detecting and removing vandalism is essential to mitigate spam and preserve the credibility and trust in OSM data.

Vandalism detection in OSM is particularly challenging due to the large dataset scale, the high number of contributors (over 7.6 million in June 2021), the variety of vandalism forms, and the lack of available annotated data.
Figure \ref{fig:intro_example} presents four real-world vandalism examples.
The vandalism forms in OSM include arbitrary deletion of map regions, creating non-existing cities, drawing texts using geometric shapes, and overwriting street names with advertisements and offensive content.
Vandalism detection methods need to consider various aspects such as the geographic context, user behavior, and content semantics to identify potentially malicious edits effectively.
The diversity of vandalism appearances constitutes a significant challenge for automated vandalism detection.

The existing literature has considered OSM vandalism previously, but only a few automated approaches for vandalism detection in OSM exist.
An early approach proposed in \cite{ijgi1030315} adopts a rule-based method to identify suspicious edits.
However, configuring the rules manually is tedious and error-prone.
In \cite{ijgi9090504}, the authors proposed a random forest-based method that detects vandalized buildings. This approach is limited to the building domain and does not detect vandalism on other OSM objects. 
Furthermore, due to the shortage of benchmark datasets, existing studies typically utilize synthetic data and lack evaluation in real-world settings.

In this paper, we present the \approach{} (\underline{O}penStreetMap \underline{V}andal\underline{i}sm \underline{D}etection) model - a novel supervised machine learning approach to detect vandalism in OSM effectively.
We propose a novel neural network architecture that adopts multi-head attention to select the most relevant edits within an individual changeset, i.e., a set of edits performed by a user within one session.
Furthermore, we propose an original feature set that captures different aspects of OSM vandalism, such as user experience and contribution content.
We train and evaluate \approach on real-world vandalism occurrences in OSM, manually identified by the OSM community.

To enable the training of supervised machine learning models, we create a new dataset, OSM-Reverts, by extracting and analyzing reverted entries from the OSM edit history. OSM-Reverts includes over 18 thousand real-world vandalism examples created by over eight thousand users during a 5-years-period on a world scale.

Although reverts indicating vandalism are available in OSM, 
identifying specific geographic entities affected by vandalism from reverts is not trivial as OSM does not specify which exact changeset is being corrected by the revert as well as due to OSM's large scale.
To tackle this problem, we developed a dedicated procedure extracting vandalism occurrences accurately at scale.

In summary, the main contributions of this paper are as follows:
\begin{itemize}
\item We present \approach{} -- a novel supervised attention-based method for vandalism detection in OpenStreetMap. 
\item We propose a set of original features that capture changeset, user and edit information to detect vandalism effectively.
\item We create OSM-Reverts, the first large open dataset capturing real-world vandalism from the OSM edit history. 
\end{itemize}

Our evaluation results on two real-world datasets demonstrate that \approach outperforms existing approaches by 8.14 percent points in F1 score and 5.41 percent points in terms of accuracy on average.
In this paper, we focus on vandalism detection in OSM.
The approach is transferrable to other collaborative knowledge bases in the geographic domain, such as geographic entities in Wikidata, providing the proposed features and annotated vandalism examples.

\section{Background \& Problem Definition}
\label{sec:problem}

%
OpenStreetMap categorizes the geographic objects into three types. 
\emph{Nodes} represent geographic points (e.g., mountain peaks) with the position specified by latitude and longitude.
\emph{Ways} represent lines (e.g., roads) composed of a node sequence. 
\emph{Relations} are composed of nodes and ways and describe more complex objects, e.g., national borders.
%
%
An OSM object may exhibit an arbitrary number of \emph{tags}, i.e., key-value pairs, describing the object semantics. 
For instance, the tag \texttt{place=city} indicates that the OSM object represents a city.

\begin{figure*}
    \centering
    \includegraphics[width=0.79\textwidth]{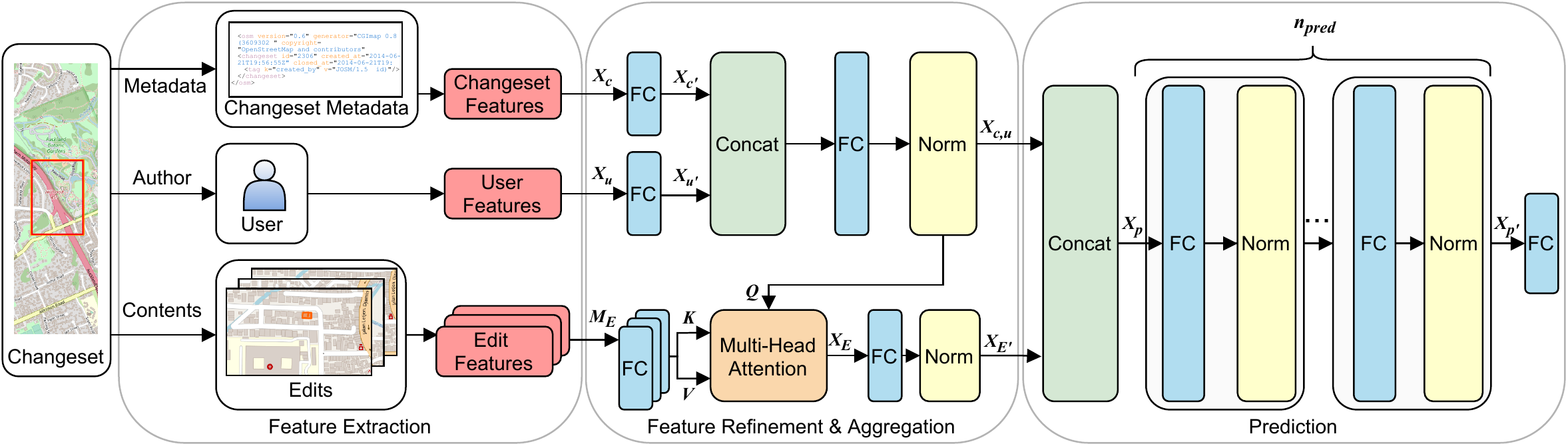}
    \caption{\approach{} model architecture. The changeset and user feature refinement part, as well as the prediction layers, are composed of fully connected (FC), normalization (Norm), and concatenation (Concat) layers. Multi-head attention layers aggregate the features from multiple edits in a changeset into a single feature vector. We describe changeset features ($X_c$) in Section \ref{sec:x_c}, user features ($X_u$) in Section \ref{sec:x_u}, and edit features ($M_e$) in Section \ref{sec:x_e}.
    Map images: \textcopyright OpenStreetMap contributors, ODbL}
    \label{fig:model_architecture}
    \Description[Neural network architecture diagram]{A neural network architecture diagram with an OSM changeset as input and a vandalism label as output.}
\end{figure*}

Formally, an \emph{OSM object} is defined as $o = \langle id, type, loc, tags, ver \rangle$.
$id$ is an object identifier.
$type \in \{\texttt{Node}, ~\texttt{Way}, ~\texttt{Relation} \}$ is the object type.
$loc$ is the geographic location of the object. The location can either be a point (Node), a line (Way), or a set of points and lines (Relation).
$tags$ is a set of tags describing object characteristics. Each tag $\langle k,v \rangle \in tags$ is represented as a key-value pair with the key $k$ and a value $v$.
$ver$ is the object version number corresponding to the number of the revisions of $o$.
An OSM object $o$ can be distinguished by its identifier $o.id$ together with its type $o.type$.

OSM allows for updates in the form of edits of the OSM objects. 
An edit can either \emph{create} new or \emph{modify} or \emph{delete} existing objects.
More formally, an \emph{edit} is defined as $e=\langle o, op, ver, t\rangle $, where: 
$o$ is an OpenStreetMap object, 
$op \in \{\texttt{create}, \texttt{modify}, \texttt{delete}\}$ is the operation performed on the object $o$, 
$ver$ is the new version number ($o.ver+1$) after the edit, 
and $t$ is the edit time. 

Edits are submitted to OSM as \emph{changesets}. 
Changesets bundle multiple edits created by a single user during a short time period. 
A \emph{changeset} is defined as $c=\langle E, t, u, co \rangle$, where:
$E$ is a set of OSM edits that belong to the changeset,
$t$ is the changeset commit time,
$u$ is the user who committed the changeset,
and $co$ is a comment describing the changeset contents.
We denote the set of all changesets by $C$.

In this paper, we target the problem of identifying vandalism changesets in OSM, i.e., the changesets containing wrong or prohibited (e.g., discriminating or offensive) content.
We follow established vandalism detection definitions from other domains, i.e., Wikidata \cite{10.1145/2983323.2983740}, and model vandalism detection in OSM as a binary classification problem.
We leave the fine-grained discrimination between various vandalism types for future work.
We assume that an adversary has all OSM user capabilities, i.e., she/he can freely submit changesets to OSM.
Currently, OSM does not implement automatic mechanisms for vandalism detection. Therefore, we do not assume that an adversary takes any specific measures to overcome the vandalism detection or creates false positive vandalism entries to mislead such mechanisms.
Formally, \emph{vandalism detection} is the task of identifying changesets that constitute vandalism by either deleting correct information or adding wrong or prohibited information. We aim to learn a function $\hat{y}:C \mapsto \{\text{True}, \text{False}\}$ that assigns binary vandalism labels to changesets.

\section{The \approach{} Model}
\label{sec:approach}
This section presents the \approach{} (\underline{O}penStreetMap \underline{V}andal\underline{i}sm \underline{D}etection) model.
\approach{} is a supervised binary classification model that discriminates between regular and vandalism OSM changesets. 
The model consists of a supervised artificial neural network, including three main components: Feature Extraction, Feature Refinement \& Aggregation, and Prediction.
Figure \ref{fig:model_architecture} provides an overview of the \approach{} model architecture.
We adopt features in three categories. 
First, changeset features capture meta-information of the individual changesets, e.g., the editor software.
Second, user features provide information regarding previous editing activities of the changeset author, e.g., the number of prior contributions.
Third, edit features encode the individual changes within the changeset, e.g., if an object was added, modified, or deleted.
Since a single changeset may consist of multiple edits, \approach{} relies on a multi-head attention mechanism to aggregate the edits and identify information relevant for vandalism detection. 
Finally, the prediction layers integrate the features and facilitate vandalism detection.

\subsection{Feature Extraction}
\label{sec:feature-extraction}

This section describes the extraction of the individual features.
\subsubsection{Changeset Features}
\label{sec:x_c}
The changeset feature vector $X_{c}$ provides information regarding the changeset metadata.
For a changeset $c$, the feature vector consists of the following features.

\textbf{No. creates} \cite{glovecnn}, \textbf{modifications} \cite{glovecnn}, \textbf{deletes} \cite{glovecnn}, \textbf{edits} \cite{glovecnn}.
We capture the changeset size by the number of created, modified, and deleted objects and the total number of edits in the changeset $|c.E|$. 
Atypical patterns, e.g., numerous deletions, can indicate vandalism.

\textbf{Min/max latitude/longitude, bounding box size.}
We capture the changeset's geographic extent by considering the minimum and maximum latitude and longitude among the changeset entries. 
We also include the size of the overall geographic bounding box of the changeset.
A large geographic extent may indicate vandalism.

\textbf{Editor application.}
Several editor applications can create OSM changesets. 
Basic editors are easy to use and, therefore, more likely to be used for vandalism. 
We include the editor application as a categorical feature using 1-hot encoding.

\textbf{Has imagery used.}
OSM contributors may specify whether they used aerial images for a changeset, which intuitively can make it more trustworthy.
We include this information as a binary variable. %

\textbf{Comment length} \cite{10.1145/2983323.2983740}.
Contributors can provide a comment $c.co$ to document the changeset.
Intuitively, a long description may indicate trustworthy changes. 
We use the number of characters in a comment as a feature.

Finally, we concatenate all changeset features to obtain the changeset feature vector $X_{c} \in \mathbb{R}^{d_{c}}$. $d_{c}$ denotes the dimension of $X_{c}$.

\subsubsection{User Features}
\label{sec:x_u}
We utilize user features to capture the previous activity of the changeset author $c.u$, as a more experienced user may be more trustworthy than a new user.
Given a changeset $c$ and its author $c.u$, we denote the user feature vector by $X_u$.

\textbf{No. past creates} \cite{ijgi1030315}\textbf{, past modifications, past deletes}.
User experience plays a vital role in quantifying user credibility \cite{ijgi1030315}. 
We quantify user experience as the number of previously created, modified, and deleted objects and add these numbers as features.

\textbf{No. contributions} \cite{10.1145/2983323.2983740, ijgi9090504}.
We count the overall number of objects contributed by the user and include this number as a feature.

\textbf{No. top-12 keys used} \cite{ijgi1030315}.
Following \cite{ijgi1030315}, we use the top-12 most frequent OSM keys and determine how often the user added one of these keys to an entry.
As of May 2020, the top-12 most frequent keys are \textit{building, source, highway, name, natural, surface, landuse, power, waterway, amenity, service, } and \textit{oneway}. 
%
%
An absence of the top-12 most essential keys in the user history might indicate harmful behavior.
We include the number of top-12 keys previously utilized by the user as a feature.

\textbf{Account creation date} \cite{10.1145/2983323.2983740}, \textbf{no. active weeks} \cite{ijgi9090504}.
To quantify the temporal scope of user experience, we consider the timestamp of the user account creation and the number of weeks in which the user has contributed at least one changeset.

We concatenate all features to obtain the user feature vector $X_{u} \in \mathbb{R}^{d_u}$, where $d_{u}$ denotes the dimension of $X_{u}$.

\subsubsection{Edit Features}
\label{sec:x_e}
The edit features capture information regarding the individual edits contained in a changeset. 
As a single changeset $c$ may contain several edits, we first extract the features for every edit $e \in c.E$.
We extract the following features from each edit $e$: 

\textbf{Object type} \cite{ijgi1030315}, \textbf{edit operation} \cite{ijgi1030315}.
The object type $e.o.type \in \{\texttt{Node}, \texttt{Way},  \texttt{Relation}\}$ and the edit operation $e.op \in \{\texttt{create}, \linebreak \texttt{modify}, \texttt{delete} \}$ provide basic information about the type of the edited object and the editing operation.
Some object types might be easier to vandalize. 
For example, it is easier to move the single node representing the South Pole than the complex relation representing Antarctica.
We use one-hot encoding to represent both features.

\textbf{Object version number, no. previous authors} \cite{10.1145/2983323.2983740, ijgi9090504}.
If the edit changes an existing object, we capture the object edit history by considering the version number $e.ver$ and the number of distinct previous authors.
A high version number might indicate controversial objects that are a subject of so-called edit wars\footnote{\url{https://wiki.openstreetmap.org/wiki/Disputes}}.
For instance, the country affiliation of some regions might be controversial\footnote{\url{https://wiki.openstreetmap.org/wiki/Disputed_territories}}.

\textbf{Time to the previous version} \cite{ijgi1030315, ijgi9090504}.
We measure the time between the current and the last object versions as their timestamp difference.
For new objects, we set this feature to zero.

\textbf{No. tags} \cite{ijgi9090504}.
Tags provide semantic information about OSM objects.
A high number of tags may indicate an established object.
Therefore, we include the total number of tags $|e.o.tags|$, the number of added tags, and the number of tags deleted in the edit as features.

\textbf{No valid tags} \cite{ijgi1030315}, \textbf{no. previous valid tags} \cite{ijgi1030315}.
The OpenStreetMap Wiki provides a description of established key-value pairs as so-called map feature list\footnote{\url{https://wiki.openstreetmap.org/wiki/Map_features}}.
Following \cite{ijgi1030315}, we count the number of tags that appear in the map feature list as the number of valid tags. 
We assume that using a valid key-value combination indicates proper editing behavior.
Likewise, we determine the number of valid tags in the previous version of the edited object if a previous version exists.
Otherwise, we set this feature to zero.

\textbf{Name changed.}
Vandalizing object names is an effective way to create visible fake information in OSM. 
We create a binary feature indicating whether the object name has changed. 
For unnamed or new objects, we set this feature to 0.

For each edit $e \in c.E$, we concatenate the features into the edit feature vector $X_{e} \in \mathbb{R}^{d_{e}}$, where $d_{e}$ denotes the dimension of $X_{e}$.
Further, we aggregate the individual features of all edits in the same changeset.
We combine the feature vectors $X_e$ for each $e \in c.E$ into the edit feature matrix $M_e \in \mathbb{R}^{d_e \times |c.E|}$.

\subsubsection{Limitations \& Adversarial Robustness}
Limitations in vandalism detection arise from the robustness of the proposed feature set in adversarial settings.
Bypassing our system requires manipulating the above-described features.
The manipulation of \emph{user features}, e.g., no. active weeks, requires a high effort, such as creating and maintaining a fake account for a long time. 
These features have a substantial impact on the model performance, as we will discuss later in Section \ref{sec:ablation}.
The attacking of other features, such as \emph{changeset and edit features}, e.g., the no. valid tags, is more straightforward but requires knowledge of our model.
Since \approach relies on a combination of user, changeset, and edit features, manipulating single features is not sufficient to bypass our system.


\subsection{Feature Refinement \& Aggregation}
\label{sec:feature-refinement}

In this step, we refine the changeset and user features and aggregate the edit features to obtain a feature vector for each changeset.

We refine the changeset and user feature vectors by passing them to the fully connected layers $X_{c'} = FC_{d_h}(X_{c})$ and  $X_{u'} = FC_{d_h}(X_u)$ with:
$
    FC_{d_h}(X_i) = ReLU(X_{i}W_{i} + b_{i}),
$
where $W_{i} \in \mathbb{R}^{d_i \times d_{h}}$ denotes a weight matrix, $b_{i} \in \mathbb{R}^{d_h}$ a bias vector,  $d_h$ is the hidden layer size, and $ReLU$ denotes the Rectified Linear Unit activation function \cite{Goodfellow-et-al-2016}.
We concatenate the changeset and user features and apply a fully connected layer with normalization:
$
    X_{c,u} = norm(FC([X_{c'}, X_{u'}])).
$
$norm(\cdot)$ denotes layer normalization \cite{ba2016layer} that scales the layer output based on the mean and the standard deviation of neuron activations.

\approach selects the edits most relevant to identify vandalism in the corresponding changeset using an attention mechanism.
First, we apply the same fully connected layer to each edit $M_{e'} = FC_{d_h}(M_e)$ to obtain the refined edit features $M_{e'}$.
Then, to aggregate the features of the individual edits into a single feature vector, we adopt the multi-head attention mechanism, initially proposed by \cite{NIPS2017_3f5ee243}. 
Intuitively, the multi-head attention mechanism computes a weighted sum of the edit features, where the model learns the so-called attention weights representing the importance of specific edits.
%

Formally, the attention mechanism distinguishes between a \emph{query} $Q$, \emph{keys} $K$ and \emph{values} $V$.
Attention selects the most relevant values $V$ for the query $Q$ based on the similarity between $Q$ and the keys $K$.
As we aim at selecting the edits most relevant to identify vandalism in the corresponding changeset,
we represent the refined changeset and user features as the query in the attention model
and the refined edit features as keys and values:
$Q = X_{c,u}$, 
$K = M_{e'}$, and
$V = M_{e'}$.
The \textit{Attention} function is defined as:
$
    \textit{Attention}(Q,K,V) = \text{softmax}(\frac{QK^T}{\sqrt{d_k}})V,
$
where $Q$ denotes a query vector, $K$ a key matrix, $V$ a value matrix, and $d_k$ is the dimension of one row (one key) of the key matrix.
The term $QK^T$ computes the similarity between the query vector $Q$ and the individual keys in the key matrix $K$.
Then, the softmax function transforms the similarities to a probability distribution representing the attention weights.
The scaling factor $\sqrt{d_k}$ prevents the softmax from having extremely small gradients during back propagation \cite{NIPS2017_3f5ee243}.
Finally, the multiplication of the attention weights with the value matrix $V$ yields the weighted sum of the values.

\textit{Multi-Head} attention extends attention by using multiple ($n_{head}$) attention heads. 
Each head learns to focus on different edit feature combinations, e.g., the object type and the tags' semantic description.
Formally, each head computes its own attention \linebreak function:
$ \textit{Multi-Head}(Q,K,V) = [\text{head}_1, ..., \text{head}_{n_{head}}]W^o, $
$
    \text{head}_i = \textit{Attention}(QW_i^Q, KW_i^K, VW_i^V),
$
with the projection matrices 
$W_i^Q \in \mathbb{R}^{d_h \times d_h}$,
$W_i^K \in \mathbb{R}^{d_h \times d_h}$,
$W_i^V \in \mathbb{R}^{d_h \times d_h}$, and
$W^O \in \mathbb{R}^{n_{head} \cdot d_h \times d_h}$.
We compute an aggregated edit feature vector using multi-head attention: $X_E = \textit{Multi-Head}(X_{c,u}, M_{e'}, M_{e'})$.
Finally, we refine the edit feature vector using a fully connected layer with the layer normalization $X_{E'} = norm(FC_{d_h}(X_E))$.

Some changesets, e.g., automatic imports, may contain a high number of edits (up to 50,000 in our datasets) such that the contribution of an individual edit is negligible.
Therefore, we introduce an upper threshold $th_{e,max}$ for the maximum number of edits within a changeset. 
If the number of edits exceeds $th_{e,max}$, we set $X_{E'}=0$ and rely on the user and changeset features.

\subsection{Prediction}
\label{sec:prediction-layers}

We facilitate the detection of vandalism changesets by combining the refined changeset and user features with the aggregated edit features into a single feature vector $X_p = [X_{c,u}, X_{E'}]$.
We repeat fully connected layers with layer normalization $n_{pred}$ times and use a final fully connected layer with a single output dimension and sigmoid activation function to make predictions of the binary vandalism label:
$
    X_{p'} = norm(FC_{d_h}(X_p))^{n_{pred}},
$
and
$
    y_{pred} = sigmoid(X_{p'} W_{p'}\ + b_{p'}),
$
with $W_{p'} \in \mathbb{R}^{1 \times d_h}$ and $b_{p'} \in \mathbb{R}$.
We determine the prediction function $\hat{y}$ by considering a changeset as vandalism if $y_{pred}$ exceeds the classification threshold: $y_{pred} > th_{class}$.
We use an established threshold for the sigmoid function $th_{class}=0.5$ \cite{Goodfellow-et-al-2016}.
We investigate the influence of $th_{class}$ on the precision and recall later in the evaluation in Section \ref{sec:precision-recall}.

\section{Datasets}
\label{sec:datasets}

We construct a novel ground truth dataset (``OSM-Reverts'') including over 9 thousand real-world vandalism incidents extracted from the world-scale OSM edit history.
We add non-vandalism changesets resulting in a dataset of over 18 thousand training examples allowing for training of supervised machine learning models.
We make OSM-Reverts available as open data to facilitate reproducibility and further research.\footnote{The OSM-Reverts dataset is available at: \url{https://github.com/NicolasTe/Ovid}.}
Furthermore, we consider a second smaller dataset (``OSM-Manual'') for our experiments.
Table \ref{tab:datasets} summarizes selected dataset statistics.

\textbf{OSM-Reverts.} 
We create a ground truth dataset by considering the changesets reverting vandalism in the OSM history from 2014 to 2019.
The correctness of the ground truth is essential for training of supervised models.
Therefore, we aim at high precision in the ground truth extraction process. 

Determining specific vandalism changesets repaired by the reverts is not trivial.
Reverts are structured as changesets and only provide a textual description of the corrections they make.
Whereas some reverts mention the corresponding changesets in the comment explicitly, 
others simply delete, update or insert geographic objects to repair the vandalism. 
The revert comments are highly heterogeneous and do not always specify the vandalism changeset.
The extraction process consists of the following steps:
First, we extract the revert changesets that fix vandalism changesets.
We only consider changesets that mention ``vandalism'' in their comments.
Second, we determine the vandalism changesets corrected by the revert.
If a revert changeset explicitly mentions a specific changeset ID, we consider the mentioned changeset as vandalism.
Otherwise, we review the objects that are the subject of the revert.
If the revert deletes an object and only one user contributed to the object, we consider changesets contributing to this object to be vandalism.
It is hard to attribute a revert to a specific changeset in other cases.
As we aim at high precision, we do not include such underspecified cases in OSM-Reverts. 

To create negative examples (i.e., changesets that do not represent vandalism), we remove the identified vandalism changesets and the reverts from the OSM changeset history.
We obtain negative examples by randomly sampling changesets from the filtered OSM history.
We randomly sample the same number of changesets as the vandalism changesets from the reduced changeset history to create negative examples and obtain a balanced dataset.
As a result, we obtain a dataset with 18,276 training examples.

Possible limitations in the ground truth quality can include the occasional occurrence of false positive and false negative examples.
\emph{False positives}, i.e. legitimate changesets labelled as vandalism, can result from malicious reverts. 
Following our adversarial model described in Section \ref{sec:problem}, we assume that malicious reverts are currently not part of our dataset. 
\emph{False negatives} are changesets containing vandalism but not labelled as such. 
Following prior studies on vandalism on crowdsourced knowledge bases \cite{DBLP:conf/sigir/HeindorfPSE15},
the overall fraction of the vandalism changesets is small compared to all OSM changesets.
Therefore, we assume that the overall fraction of false negatives in our datasets is neglectable.

We split the dataset into training (70\%), validation (10\%), and test (20\%) sets.
We ensure that the training, validation, and test sets are disjunct concerning OSM users to avoid bias towards individual OSM users.
We train the models on the training and validation set and evaluate the results on the test set.

\textbf{OSM-Manual.} 
In 2018, the OSM community manually identified approximately one thousand vandalism incidents, including spam and forbidden imports.\footnote{OSM-Manual data: \url{https://github.com/jremillard/osm-changeset-classification}}
We use these changesets as positive examples. 
We create negative examples by random sampling changesets from the filtered OSM history.
In total, the OSM-Manual dataset includes 2,018 examples. 
The number of examples in the OSM-Manual dataset is too small to train supervised models.
Thus, we use OSM-Manual only as a test set. 
We use OSM-Manual for evaluation by training the models on the entire OSM-Reverts dataset and then using OSM-Manual as a test set.

\begin{table}
    \caption{Dataset statistics for OSM-Reverts and OSM-Manual}
    \resizebox{0.45\textwidth}{!}{
    \begin{tabular}{lrr}
    \toprule
     Dataset property & OSM-Reverts & OSM-Manual \\
    
    \midrule
    No.  changesets    & 18,276  & 2,018 \\
    No. distinct users & 8,768 & 1,686 \\
    Median create operations per changeset & 5 &  3\\
    Median modify operations per changeset & 1 &  1\\
    Median delete operations per changeset & 1 &  1\\
    Median nodes per changeset & 6 &  4\\
    Median ways per changeset & 2 &  2\\
    Median relations per changeset & 1 &  1\\
     
    Median edits per changeset & 10 &  8\\
    Median edits/vandalism changeset & 10 & 4 \\
    Median edits/negative changeset & 10 & 11 \\
    Timespan & 2014-2019 & 2014-2019\\
    \bottomrule
         
    \end{tabular}}
    \label{tab:datasets}
\end{table}
    
\newcommand{\random}{\textsc{Random}\xspace}
\newcommand{\patrol}{\textsc{OSMPatrol}\xspace}
\newcommand{\watchman}{\textsc{OSMWatchman}\xspace}
\newcommand{\WDVD}{\textsc{WDVD}\xspace}
\newcommand{\Glove}{\textsc{GloVe+CNN}\xspace}

\section{Evaluation Setup}
\label{sec:setup}
This section describes the baselines, metrics, and hyperparameter optimization used in the evaluation.

\subsection{Baselines}
\label{sec:baseline}
We compare our model with the following baselines:

\random. This na\"ive baseline chooses a random vandalism label.

\patrol. This model is an early rule-based approach to detect OSM vandalism at the edit level \cite{ijgi1030315}. 
\patrol computes a vandalism score for each edit based on a rule combination considering user and edit features, e.g., the object version number.
We consider a changeset $c$ as vandalism if the baseline detects at least one vandalism edit $e \in c.E$ in this changeset.
We apply an exhaustive grid search to find the optimal thresholds.

\watchman. This model was recently proposed to detect vandalism on buildings in OSM \cite{ijgi9090504}. 
\watchman uses a random forest classifier that utilizes content features (e.g., number of tags), context features (e.g., time to the previous version), and user features (e.g., number of contributions).

\WDVD. The Wikidata Vandalism Model was proposed to detect vandalism in the Wikidata knowledge graph \cite{10.1145/2983323.2983740}. This baseline uses a random forest classifier with text-based features, e.g., the ratio of upper case letters in comments, and user features.  
We use all features applicable to OSM.

\Glove. This baseline \cite{glovecnn} transforms OSM changesets into pseudo-natural language sentences describing the changeset contents. 
It then uses pre-trained GloVe word embeddings \cite{pennington2014glove} as an input for a convolutional neural network.

We optimize the hyperparameters of \watchman, \WDVD, and \Glove using random search.

\subsection{Metrics}
\label{sec:metrics}

We use the following metrics for the performance evaluation:

\noindent\textbf{Precision.} The fraction of correctly classified vandalism instances among all instances classified as vandalism.

\noindent\textbf{Recall.} The fraction of correctly classified vandalism instances among all vandalism instances.

\noindent\textbf{F1 score.} The harmonic mean of recall and precision.

\noindent\textbf{Accuracy.} The fraction of correctly classified instances. 

\begin{table}
    \caption{Hyperparameter search space of \approach}
    \small
    \begin{tabular}{lll}
        \toprule
         Parameter & Description & Search Space \\
         \midrule
         $th_{e,max}$ & \makecell[l]{Maximum number of edits\\ per changeset}   & $\{10, 20, 30\}$ \\
         $n_{pred}$ & Number of prediction layers & $\{1,2,3,4,5\}$\\
         $n_{head}$ & Number of attention heads & $\{5, 10, 15, 20\}$\\
         $d_h$ & Hidden layer size & $\{12, 24, 36, 48\}$\\
         $\delta$ & Dropout rate & $\{0.4, 0.5, 0.6, 0.7 \}$\\
         $\lambda$ & Regularization weight & $\{0.005, 0.01, 0.02\}$\\
         \bottomrule
    \end{tabular}
    \label{tab:hyperparams}
\end{table}

\begin{table*}
\small
\caption{Vandalism detection performance regarding precision, recall, F1 score and accuracy [\%]. Best scores are marked bold.}
\begin{tabular}{l@{\quad}cccc@{\hskip 1em}cccc@{\hskip 1em}cccc}
\toprule
\multirow{2}{*}{\raisebox{-\heavyrulewidth}{Approach}} &               \multicolumn{4}{l}{OSM-Reverts} 
	& \multicolumn{4}{l}{OSM-Manual}
		& \multicolumn{4}{l}{Average}\\
\cmidrule(l{0pt}r{6pt}){2-5} \cmidrule(l{0pt}r{6pt}){6-9} \cmidrule(l{0pt}r{6pt}){10-13}  
    & Precision & Recall & F1 & Accuracy
    & Precision & Recall & F1 & Accuracy
    &  Precision & Recall & F1 & Accuracy  \\  
\midrule

\random  & 49.89 & 50.27  & 50.08 & 49.92
            & 49.46 & 49.95 & 49.70 & 49.45
            & 49.68 & 50.11 & 49.89 & 49.69\\
\patrol   & 53.94 &  \textbf{96.29} & 69.15 & 57.06
         & 52.20 & \textbf{91.77} & 66.55 &  53.87
         & 53.07 & \textbf{94.03} & 67.85 & 55.47 \\
\watchman   & 77.60 & 70.74  & 74.01 & 75.18
            & 57.75 &  32.11 & 41.27 & 54.31
            & 67.68 & 51.43 & 57.64 & 64.74 \\
\WDVD   &  \textbf{81.52} & 64.79  & 72.20 & 75.07
        & 74.88 & 46.98 & 57.73 & 65.61
        & 78.20 & 55.88 & 64.97 & 70.34\\
\Glove   & 81.46 & 72.93 & 76.96 & 78.18  
        & \textbf{82.16} & 19.62 & 31.68 & 57.68
        & \textbf{81.81} & 46.27 & 54.32 & 67.93\\

\midrule

\textsc{\approach} & 80.35 & 83.02  & \textbf{81.66} & \textbf{81.37}
                    & 69.86 & 70.76 & \textbf{70.31} & \textbf{70.12}
                    & 75.11 & 76.89 & \textbf{75.99} & \textbf{75.75}\\
     \bottomrule
\end{tabular}
\label{tab:results}

\end{table*}

\noindent We consider F1 score and accuracy as most relevant for this study.

\subsection{Hyperparameter Tuning \& Training}
\label{sec:tuning}
We normalize all features and optimize hyperparameters of \approach using random search.
Table \ref{tab:hyperparams} summarizes the hyperparameter space.
We train \approach using the ADAM optimizer \cite{DBLP:journals/corr/KingmaB14} and dropout layers.
We use 100 epochs and apply the early stopping strategy \cite{Goodfellow-et-al-2016}.

\newcommand{\approachsc}{\approach}

\section{Evaluation}
\label{sec:evaluation}

The evaluation aims to assess the effectiveness of the proposed \approach{} approach for vandalism detection.
Furthermore, we analyze the contribution of the feature categories in an ablation study and investigate the capability to customize our approach concerning the trade-off between precision and recall.  

\subsection{Vandalism Detection Performance}
\label{sec:eval_performance}

Table \ref{tab:results} summarizes the overall vandalism detection performance of the \random, \patrol, \watchman, \WDVD, and  \Glove baselines as well as our proposed \approachsc approach.

Overall, we observe that in terms of F1 score and accuracy, \approachsc achieves the best performance on both datasets.
On average, \approachsc achieves improvements of  8.14 percent points F1 score and 5.41 percent points accuracy compared to the best performing baseline.

Exclusively optimizing only one of precision and recall is insufficient for effective vandalism detection. 
Consider the \patrol baseline that only achieves precision scores close to 50\% but recall scores higher than 90\%. 
The scores reveal that \patrol assigns almost all changesets to the vandalism class, resulting in low average accuracy of 55.47\%.
Although recall and precision help understanding the model behavior, they should only be consulted jointly with accuracy in our settings.
The low accuracy of the baseline indicates that supervised machine learning models like \approachsc can better detect vandalism than rule-based \patrol.

\WDVD achieves a high precision but only reaches low recall.
\WDVD mainly relies on user features. 
The high precision indicates that user features can effectively identify a fraction of the malicious changesets.
However, the low recall score indicates that user features are insufficient to capture all vandalism incidents.
Low recall means that many vandalism cases will remain undetected when using this baseline. 
In contrast, \approach that considers user, changeset, and edit features, achieves 83.02\% recall on OSM-Reverts.

\Glove achieves the best precision on OSM-Manual but only achieves a recall score of 19.62\% resulting in a relatively low F1 score of 31.68\%.
\Glove does not consider user information.
Comparing \Glove to \WDVD, we observe that \WDVD achieves higher recall and F1 scores on OSM-Manual than \Glove.
This result indicates the importance of the user features for the OSM-Manual dataset.
\approachsc that combines user and content features achieves a high recall (70.76\%) on OSM-Manual while maintaining a comparably high precision (69.86\%).

\watchman shows a moderate performance considering all metrics on OSM-Reverts but fails to maintain the performance level on OSM-Manual.
\watchman uses a combination of user and content features.
The low F1 score on OSM-Manual of 41.27\% indicates that the feature set of \watchman does not generalize.

Comparing the performance across the datasets, we generally observe higher scores on OSM-Reverts than on OSM-Manual.
As OSM-Manual is too small to train supervised machine learning models, we trained all models on OSM-Reverts for both datasets, as described in Section \ref{sec:datasets}. 
The difference in the performance indicates that the datasets exhibit slightly different underlying distributions.
We observe the best performance when we train and evaluate the models on the train and test datasets with the same distribution, i.e., OSM-Reverts.
The difference in distributions, also indicated by Table \ref{tab:datasets}, can result from the possible presence of false negatives and false positives in OSM-Reverts as discussed in Section \ref{sec:datasets}.
These false examples in the training data can potentially lead to the difference in performance between OSM-Reverts and OSM-Manual.
However, \approachsc's good performance on OSM-Manual indicates the 
usefulness of OSM-Reverts as a training dataset and 
\approachsc's ability to generalize to unseen data.

The \Glove baseline, achieving the second-best F1 score on OSM-Reverts, fails to generalize to OSM-Manual and only achieves an F1 score of 31.68\%.
In contrast, \approachsc achieves higher than 70\% F1 score and accuracy.
This observation indicates that \approach's features and architecture better generalize to unseen data than the baselines.

\newcommand{\ablChange}{$\textsc{\approach}_{-Changeset}$\xspace}
\newcommand{\ablUser}{$\textsc{\approach}_{-User}$\xspace}
\newcommand{\ablEdits}{$\textsc{\approach}_{-Edits}$\xspace}
\newcommand{\ablChangeEdits}{$\textsc{\approach}_{-Changeset,Edits}$\xspace}
\newcommand{\ablUserEdits}{$\textsc{\approach}_{-User,Edits}$\xspace}

\begin{table}
    \small
    \caption{Vandalism detection performance of \approach when removing individual components. Best scores are bold.}
    \begin{subtable}{0.45\textwidth}
    \centering
    \caption{OSM-Reverts}
    \begin{tabular}{lcccc}
    \toprule
    Model & Precision & Recall & F1 & Accuracy \\
    \midrule
    \textsc{\approach} &  \textbf{80.35} & 83.02 & \textbf{81.66} & \textbf{81.37}\\
    \midrule
    \ablChange & 78.42 & 81.33 & 79.85 & 79.49\\
    \ablUser &  67.46 & \textbf{86.24} & 75.71 & 72.34\\
    \ablEdits & 78.68 & 78.98 & 78.83 & 78.81\\
    \ablChangeEdits & 75.69 & 81.22 & 78.36 & 77.58\\
    \ablUserEdits & 59.54 & 72.38 & 65.34 & 61.62\\
    
    \bottomrule
    \end{tabular}
    \label{tab:ablation_reverts}
    \end{subtable}
    \begin{subtable}{0.45\textwidth}
    \centering
    \caption{OSM-Manual}
    \begin{tabular}{lcccc}
    \toprule
    Model & Precision & Recall & F1 & Accuracy \\
    \midrule
    \textsc{\approach} & 69.86 & \textbf{70.76} & 70.31 & 70.12\\
    \midrule
    \ablChange & 59.03 & 49.55 & 53.88 & 57.58\\
    \ablUser  &  53.83 & 62.74 & 57.94 & 54.46\\
    \ablEdits & \textbf{73.70} & 67.49 & \textbf{70.46} & \textbf{71.70}\\
    \ablChangeEdits  & 66.94 & 55.80 & 60.86 & 64.12\\
    \ablUserEdits & 56.02 & 62.24 & 58.97 & 56.69\\

    \bottomrule
    \end{tabular}
    \label{tab:ablation_manual}
    \end{subtable}
    \label{tab:ablation}
\end{table}

\subsection{Ablation Study}
\label{sec:ablation}

We conduct an ablation study to assess the contribution of \approachsc's feature categories.
To this end, we remove individual parts of our model and measure the vandalism detection performance on the OSM-Reverts and OSM-Manual datasets.
We consider the following configurations for the ablation study:
     
\noindent     \ablChange: We remove $X_{c'}$, i.e., the changeset features and the corresponding refinement layer.
    
\noindent     \ablUser: We remove $X_{u'}$, i.e., the user features and the corresponding refinement layer.

\noindent     \ablEdits: We remove $X_{E'}$, i.e., the edit features and the corresponding multi-head attention and refinement layers.

\noindent     \ablChangeEdits: We remove $X_{c'}$ and $X_{E'}$, i.e., the changeset and edit features and the corresponding layers.

\noindent     \ablUserEdits: We remove $X_{u'}$ and $X_{E'}$, i.e., the user and edit features and the corresponding layers.

We do not remove $X_{c'}$ and $X_{u'}$ simultaneously, since the edit features aggregation component requires at least one of $X_{c'} $ or $X_{u'}$ to provide the query vectors for the multi-head attention.
Table \ref{tab:ablation_reverts} and \ref{tab:ablation_manual} present the results on OSM-Reverts and OSM-Manual.
%

On the OSM-Reverts dataset, we cannot remove any feature category without reducing the vandalism detection performance.
In other words, every feature category of our model contributes to the vandalism detection performance.
Considering configurations removing only one feature category, we observe the highest difference in accuracy for \ablUser.
This configuration leads to a slight increase in recall (3.22 percent points) but a higher decrease in precision (12.89 percent points), 
which signals that the user features are especially beneficial for precision.
For \ablEdits, we observe moderate losses on both recall and precision.
We obtain similar results for the \ablChange scores, i.e., a moderate difference in precision and recall of approximately 1.8 percent points.
The loss on all metrics indicates the general usefulness of the edit and changeset information.
Removing two feature categories simultaneously further reduces the accuracy compared to removing only one category, indicating that the categories capture different aspects of the changesets and complement each other.
For \ablUserEdits, we observe the lowest accuracy scores of 61.62\%, whereas we still obtain 77.58\% accuracy for \ablChangeEdits.
The moderate accuracy of \ablChangeEdits highlights the user features' importance, since this configuration solely relies on user information.
%

On the OSM-Manual dataset, we obtain patterns similar to OSM-Reverts for the \ablChange and \ablUser. 
For \ablUser, we observe a performance drop, especially regarding precision.
In contrast to OSM-Reverts, \ablEdits achieves an increased classification accuracy of 1.58 percent points.
As discussed in Section \ref{sec:eval_performance}, we expect that the underlying distributions of OSM-Reverts and OSM-Manual datasets can slightly differ.
In particular, the median number of edits per vandalism changeset in OSM-Reverts is ten, while this number in OSM-Manual is only four, as shown in Table \ref{tab:datasets}.
Consequently, \approachsc cannot use the edit features to their full advantage on OSM-Manual, where less edit information is available.
As a result, the edit features slightly negatively impact the vandalism detection performance on the OSM-Manual dataset. Nevertheless, \approachsc still achieves the best performance in F1 score and accuracy compared to the baselines.

In summary, the changeset and user features provide valuable contributions for the vandalism detection task. The edit features and their corresponding layers $X_{E'}$ are beneficial for the datasets with a higher number of edits per changeset, like OSM-Reverts.

\subsection{Precison/Recall Trade-off}
\label{sec:precision-recall}

\begin{figure}
    \centering
    \includegraphics[width=0.40\textwidth]{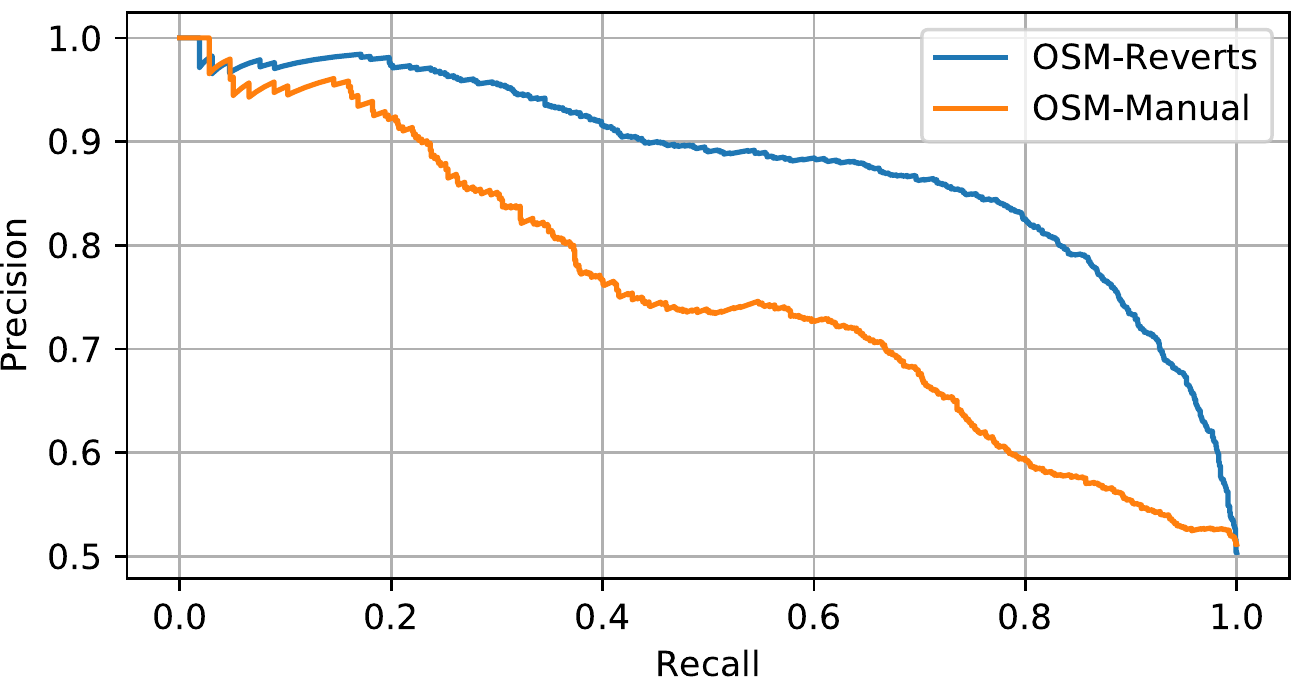}
    \caption{Precision/recall diagram of \approach.}
    \label{fig:prec_rec}
\end{figure}

Figure \ref{fig:prec_rec} presents \approachsc's precision/recall diagram created by varying the classification threshold $th_{class} \in [0,1]$ on the OSM-Reverts and OSM-Manual datasets.
At the most left, all changesets are classified as non-vandalism, such that \approach achieves 100\% precision but 0\% recall.
At the most right, all changesets are classified as vandalism, resulting in 50\% precision and 100\% recall. 
The precision does not drop to 0\% due to the class balance within the datasets.

We generally observe higher precision and recall scores on OSM-Reverts than on OSM-Manual.
On OSM-Reverts, \approachsc can obtain very high precision of 97\% at the cost of only achieving 20\% recall. 
Similarly, we obtain 92\% precision at 20\% recall for OSM-Manual.
A high-precision configuration can potentially be used to run a fully automated vandalism detection system that blocks detected changesets directly.
Lowering precision leads to a rapid increase in recall on OSM-Reverts.
At 80\% precision, \approachsc already achieves 83\% recall. 
For OSM-Manual, \approachsc maintains approximately 70\% precision at 70\% recall. 
High recall of vandalism cases may be favourable to maintain high data quality in OSM but may lead to accidental blocking of correct changesets (false positives).
A high-recall configuration of \approachsc can be used to generate vandalism candidates in a human-in-the-loop approach, 
in which vandalism candidates are verified manually by the OSM community.

\section{Related Work}
\label{sec:relatedwork}

This section discusses related work in vandalism detection in crowdsourced knowledge bases and vandalism characterization in OSM.

The existing literature investigated vandalism detection in crowdsourced knowledge bases such as OpenStreetMap in several studies.
Neis et al. early proposed \patrol \cite{ijgi1030315}, a rule-based system to detect vandalism in OpenStreetMap. %
\patrol determines a vandalism score for each edit. 
The score includes features such as user reputation, object type, and the number of established tags in the edit.
While \patrol aims at classifying individual edits, we classify entire changesets.
In this paper, we use \patrol as a baseline. 
Our experimental results confirm that \approach outperforms this baseline regarding precision, F1 score, and accuracy.

More recently, another line of research has investigated the validation of building shapes within OpenStreetMap.
Xie et al. developed a convolutional neural network that extracts building shapes from remote sensing imagery. The authors then compare the extracted shapes with shapes from OSM \cite{8924753} to validate the buildings in OSM.
The \watchman approach uses a supervised random forest model for the detection of vandalism on buildings in OSM \cite{ijgi9090504}. However, the authors evaluated \watchman on synthetically created vandalism incidents only. 
Li et al. exploited user profiles for the detection of name vandalism on OSM objects \cite{vand_statemap, vand_cikm}.
Whereas the approach for shape verification and name vandalism identification presented in \cite{8924753, vand_statemap, vand_cikm} are too specific for our experimental setting, we compare \approach to \watchman as a baseline. Our evaluation demonstrates that \approach outperforms \watchman concerning all considered metrics.

Heindorf et al. investigated the problem of vandalism detection in the Wikidata knowledge graph \cite{10.1145/2983323.2983740, DBLP:conf/sigir/HeindorfPSE15, DBLP:conf/www/HeindorfSEP19}.
They developed the \emph{Wikidata Vandalism Detection} (\WDVD) model that uses a random forest classification model together with user-based features (e.g., number of previous contributions) and text-based content features (e.g., the ratio of uppercase letters).
We compare to \WDVD as a baseline and show that our proposed \approach model outperforms \WDVD concerning F1 score and accuracy.

Another class of approaches aims at detecting vandalism of textual knowledge bases such as Wikipedia \cite{10.1007/978-3-540-78646-7_75, 10.1145/2783258.2783367}.
These approaches use Wikipedia-specific features (e.g., edits of meta-pages) and features tailored to natural language texts (e.g., the fraction of pronouns in a text).
In contrast, OSM provides object descriptions as key-value pairs that do not usually contain long natural language texts, such that this type of models does not apply to OSM data.

Previous research has investigated the characteristics of vandalism in OSM.
Antoniou et al. identified vandalism as a threat to the OSM data quality in a recent survey \cite{vgi_future}.
Quin et al. analyzed OSM user bans and further specified several threat categories such as \emph{nefariousness, obstinance, ignorance}, and \emph{mechanical problems} \cite{threads_data_quality}.
Similarly, Ballatore et al. coined the term ``carto-vandalism'' \cite{doi:10.1179/1743277414Y.0000000085}.
They categorize vandalism incidents in the types \emph{play, ideological, fantasy, artistic, industrial}, and \emph{spam} carto-vandalism.
The authors point out the potential use of automated tools such as machine learning for vandalism detection.
Mooney et al. analyzed high frequently edited objects and found so-called ``edit wars'' in OSM \cite{fi4010285}. 
Edit wars are disputes of two or more contributors in which the contributors repeatedly revert each other's contributions.
Edit wars are considered vandalism or bad-editing by the OSM community\footnote{See: \url{https://wiki.openstreetmap.org/wiki/Vandalism}}.   

These studies highlight the importance of mitigating vandalism in OpenStreetMap.
Our proposed \approach model can lower the effort required for vandalism detection in OSM in the future.

\section{Conclusion}
\label{sec:conclusion}

In this paper, we proposed the \approach (\underline{O}penStreetMap \underline{V}andal\underline{i}sm \underline{D}etection) model, a novel supervised attention-based approach for vandalism detection in OpenStreetMap. 
\approach relies on the original changeset, user, and edit features and a novel multi-head attention architecture to effectively identify vandalism changesets in OSM.
We systematically analyzed vandalism-related reverts in the OpenStreetMap history and extracted a new open ground truth dataset for vandalism detection in OSM.
Our experiments on real-world datasets demonstrate that \approach can effectively detect OSM vandalism.
\approach achieves an F1 score of 75.99\% and an accuracy of 75.75\% on average, which corresponds to 8.14 percent points increase in F1 score and a 5.41 percent point increase in accuracy compared to the best performing baselines.
In future work, we would like to build upon our vandalism detection model to create novel applications.

 \subsubsection*{Acknowledgements} This work was partially funded by DFG, German Research Foundation (``WorldKG'', 424985896), the Federal Ministry for Economic Affairs and Energy (BMWi), Germany (``d-E-mand'', 01ME19009B and ``CampaNeo'', 01MD19007B),  and the European Commission (EU H2020, ``smashHit'', grant-ID 871477).

\balance
\bibliographystyle{ACM-Reference-Format}
\bibliography{ref}

\end{document}